\documentclass[sn-mathphys-num, twocolumn]{sn-jnl}% Math and Physical Sciences Numbered Reference Style %%\documentclass[sn-aps]{sn-jnl}% American Physical Society (APS) Reference Style
\usepackage{natbib}

\RequirePackage[exponent-product = \cdot]{siunitx}
\usepackage[skip=4pt, indent=1.5em]{parskip}
\usepackage{amsmath}
\usepackage[noabbrev]{cleveref}
\crefname{figure}{Fig.}{Figs.}
\hypersetup{hidelinks=true}

\usepackage[shortcuts=abbreviations]{glossaries-extra}
\setabbreviationstyle{long-short-sc}

\usepackage{textgreek} % Greek letters in bibliography

\newabbreviation{discrete dislocation dynamics}{ddd}{discrete dislocation dynamics}
\newabbreviation{continuum dislocation dynamics}{cdd}{continuum dislocation dynamics}
\newabbreviation{micropipe}{MP}{micropipe}
\newabbreviation{TSD}{TSD}{threading screw dislocation}
\newabbreviation{TED}{TED}{threading edge dislocation}
\newabbreviation{BPD}{BPD}{basal plane dislocation}
\newabbreviation{screw dislocation}{SD}{screw dislocation}
\newabbreviation{RHS}{RHS}{right-handed screw}
\newabbreviation{LHS}{LHS}{left-handed screw}
\newabbreviation{SWXRT}{SWXRT}{whitebeam synchrotron x-ray topography}
\newabbreviation{TEM}{TEM}{transmission electron microscopy}
\newabbreviation{GA}{GA}{genetic algorithm}
\newabbreviation{DEAP}{DEAP}{distributed evolutionary algorithms in python}
\newabbreviation{PVT}{PVT}{physical vapor transport}
\newabbreviation{PCA}{PCA}{principal component analysis}
\newabbreviation{t-SNE}{t-SNE}{T-distributed stochastic neighbour embedding}
\newabbreviation{UMAP}{UMAP}{uniform manifold approximation and projection}
\newabbreviation{RMSE}{RMSE}{root mean squared error}
\newabbreviation{CNN}{CNN}{convolutional neural network}

\begin{document}
	
% \title[Deep learning and automated analysis of KOH microscropy images of 4H-SiC wafers]{Deep learning and automated analysis of KOH microscropy images of 4H-SiC wafers}
\title[Combining unsupervised and supervised learning in microscopy enables defect analysis of a full 4H-SiC wafer]
{Combining unsupervised and supervised learning in microscopy enables defect analysis of a full 4H-SiC wafer}


\author*[1]{\fnm{Binh Duong} \sur{Nguyen}}\email{bi.nguyen@fz-juelich.de}

\author[2]{\fnm{Johannes} \sur{Steiner}}

\author[2]{\fnm{Peter} \sur{Wellmann}}

\author*[1,3]{\fnm{Stefan} \sur{Sandfeld}}\email{s.sandfeld@fz-juelich.de}

\affil[1]{\orgdiv{Institute for Advanced Simulations -- Materials Data Science and Informatics (IAS-9)}, \orgname{Forschungszentrum J\"ulich GmbH}, \orgaddress{\city{J\"ulich}, \postcode{52425}, \country{Germany}}}

\affil[2]{\orgdiv{Crystal Growth Lab, Materials Department 6}, \orgname{Friedrich-Alexander University Erlangen-Nuremberg}, \orgaddress{\city{Erlangen}, \postcode{91058}, \country{Germany}}}

\affil[3]{\orgdiv{Chair of Materials Data Science and Materials Informatics, 
        Faculty 5 -- Georesources and Materials Engineering}, \orgname{RWTH Aachen University}, \orgaddress{\city{Aachen}, \postcode{52056}, \country{Germany}}}


\abstract{
    Detecting and analyzing various defect types in semiconductor materials is
    an important prerequisite for understanding the underlying mechanisms
    as well as tailoring the production processes. Analysis of 
    microscopy images that reveal defects typically requires image analysis 
    tasks such as segmentation and object detection. 
    With the permanently increasing amount of data that is produced by
    experiments, handling these tasks manually becomes more and more impossible.
    In this work, we combine various image analysis and data mining techniques 
    for creating a robust and accurate, automated image analysis pipeline. This 
    allows for extracting the type and position of all defects in a 
    microscopy image of a KOH-etched 4H-SiC wafer that was stitched together from  approximately 40,000 individual images. 
}




\maketitle

%=============================================================%%
\section{Introduction}
%=============================================================%%
Microscopy and image analysis has been an important tool for investigating
defects in materials and their microstructures. Manually detecting, e.g., 
cracks, grains or dislocations in photographs and digital images requires
expertise, experience, and -- depending on the size of the image or the 
number of features of interest -- also a significant amount of time.
Recently, high-throughput data analysis, which is concerned with handling 
huge numbers of specimens (e.g., even exceeding thousands of specimens)
or with very large regions of interest (e.g., consisting of several thousands of images)
is becoming more prominent in the era of data. Example of such approaches
can be found in the work of \citet{xiang1995combinatorial} or \citet{ludwig2019discovery} 
where combinatorical high-throughput methods were introduced to discover new materials. 
An other example is the work by \citet{castelli2012computational} who analyzed a large space of 5400 different materials to obtain 15 promising candidates for  developing new photoelectrochemical cells with improved light absorption. 
%In drug discovery, quickly identifying potential lead candidates with high accuracy would require to screen many thousand of chemical compounds \citep{malo2006statistical}. 

In the area of wide-bandgap semiconductor material, silicon carbide (SiC) is the leading candidate with a high mechanical, chemical and thermal stability. It has been shown to be highly suitable for power device applications. Therefore, efforts to produce high-quality SiC by reducing defects (especially dislocations) during the \gls{PVT} crystal growth process are important and require further optimization.
Visual inspection and analysing dislocation in semiconductors helps to
understand the formation of defects during the \gls{PVT} growth process and
ultimately ensures the wafer's quality during the production process. 
Such analysis is based on images produced by optical and scanning electron microscopy of the wafer surface or transmission 
electron microscopy techniques and X-ray topography which allow to follow single dislocation lines within the wafer. Those tasks are commonly done manually or based on classical image analysis methods that perform pixel-based operations, such as thresholding, line 
thinning, watersheding, to name but a few. These methods typically work only for
a particular image and require significant experimentation for determining suitable parameters.

In recent years, machine learning approaches have become popular in many research areas, e.g., to automate and accelerate the process of material discovery and design \citep{de2017use, correa2018accelerating, ren2018accelerated, li2020accelerated, pyzer2022accelerating, siriwardane2022generative, lyngby2022data, srinivasan2022machine, rao2022machine}, or to automate computer vision tasks such as object detection or segmentation \citep{decost2015computer, decost2017computer, roberts2019deep, horwath2020understanding, durmaz2021deep, lin2022deep}. 
%
% In addition, \citet{bian2021deep} proposes a deep neural network for high-throughput images of organoids in order to increase detection and tracking speed at high accuracy. \citet{kabiraj2020high} uses high-throughput automated codes and data-driven models to screen materials databases for predicting high Curie point material.
%
%
%
However, obtaining large enough datasets for supervised training is still a 
challenge since it is time-consuming to annotate large datasets with potentially
vast numbers of objects. Thus, ``synthetically generated data'' becomes helpful and enables the algorithm to train more effectively. 
E.g., \citet{tremblay2018training} generated synthetic objects on top of a random image background and shows that training with synthetic data performs better than with real images.
\citet{Trampert2021} created artificial grain structures and used only a few hand-labelled real microscopy images to train a \gls{CNN}. 
Similarly, \citet{Govind2024} created artificial transmission electron microscopy images and used those to train a \gls{CNN} for image segmentation tasks. 
%
% Schuette \citep{dumont2021overcoming} generates synthetic images in the medical research domain and shows that insights derived from synthetic images are similar to real images and that sharing synthetic data is preferable to real patient data in the right setting. 
% %
% By utilising synthetic datasets, Toda \citep{toda2020training} trains an instance segmentation of a deep neural network to process crop seed phenotyping real-world images and shows that this is a powerful application in the agricultural domain for reducing labour costs.
%

In this paper, we propose a combination of different data analysis techniques and deep learning methods in the material science domain for solving two main tasks:
The first task is to generalize and automatize the process of creating a dictionary pool of etch-pit images that can be used for creating artificial training data.
The second task is to use a deep learning framework to segment and count the
occurrence of three different dislocation types (\gls{BPD}, \gls{TED} and \gls{TSD}) that commonly appear as etch-pits in KOH etching microscopy images of a 4H-SiC wafer. This allows to analyze a huge number of microscopy images with high fidelity to estimate dislocation distributions of different types and thereby helps to understand mechanisms that lead to defect formation and organisation.

The paper is organised as follows: after the introduction section, in section $2$, we describe all methods that were used in our work. In section $3$, we present the results from the automated clustering process and the instance segmentation of various dislocation types. In section $4$, results are discussed, followed by the conclusion in section $5$.

%===============================================================================
\section{Materials and Methods}
%===============================================================================
The goal of this work is to analyze the dislocation content of a large SiC 
wafer of \SI{10}{\cm} in diameter. Subsequently, we start by describing the growth and preparation of 
the SiC crystal. This is then followed by introducing the imaging of the wafer 
as well as the automated machine learning pipeline for image analysis.

%-------------------------------------------------------------------------------
\subsection{Materials, specimen preparation and microscopy}
%-------------------------------------------------------------------------------
The SiC sample was cut from a crystal boule grown via the \gls{PVT} method 
utilizing a RTD-6800 diamond wire saw. The crystal was sliced parallel to the 
seed direction, resulting in samples with a 4° off-axis angle with respect to 
the $(0001)$-plane, the same as the employed seed. 

As a next step, the wafer was polished to remove the deformation zone induced 
by the sawing process. KOH platelets are heated up to 520°C. The sample is 
preheated and subsequently lowered into the melt inside a nickel sample holder 
for 7 minutes. After the etching step is completed, the sample is taken out of 
the melt, cooled down and cleaned with HCl and de-ionized water to remove any 
KOH residue. The employed setup is an in-house development of the group of one 
of the co-authors. The process is automated, spanning the preheating phase to 
the cool-down phase.  One of the two resulting pieces came into close contact 
with the sample holder, inducing the sample holder’s geometry as a pattern, as 
seen in the microscopic images. This pattern is, therefore, not indicative of 
the sample’s crystal lattice property but, instead, a result of the reduced 
exposure to the KOH melt. 
\begin{figure*}
    \centering
    \includegraphics[width=\textwidth]{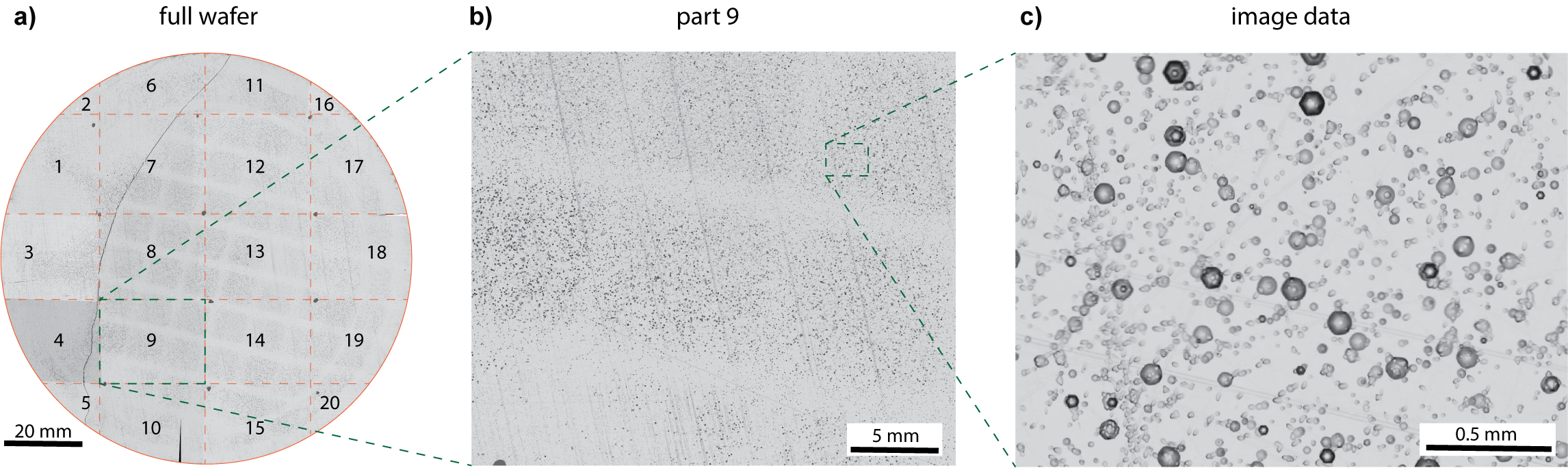}
    \caption{\label{fig:wafer}
        Micrograph of the investigated SiC waver. The magnified region shows dislocation lines piercing the surface, revealed by KOH etching.
        The whole wafer image consist of altogether $40,000$ images, one of them is shown 
        in sub-figure c).
    }
\end{figure*}

To reveal the location of dislocations the sample was etched with molten KOH. 
For this, the Si-face of a SiC wafer was cut parallel to the $(0001)$-plane and
was etched selectively by KOH etching, revealing etch pits where dislocation 
lines pierce the surface. 
Both Si-face and C-face are etched simultaneously, but only the Si-face will be considered due to its anisotropic etching nature of etch pits.
%The C-face was not chosen for etching due to much higher etching rate \citep{katsuno1999mechanism}. 
\glspl{BPD} are 
located in the basal plane, i.e., parallel to the $(0001)$-plane. Thus, they 
can be detected by KOH etching due to the present small off-axis angle which is 
4° with respect to the basal plane. 

Microscopy images of the KOH-etched sample’s Si-faces were taken utilizing a 
Zeiss Axio Imager.M1m microscope. The magnification was set to $20$ times, and 
the mapping and stitching were carried out by the accompanying Zeiss Zen 
microscopy software. The whole wafer is shown in \cref{fig:wafer}a) and is 
divided into $20$ sections for the scanning process. Each of them again consists of 
around $2000$ images with $1292\times968$ pixels (see \cref{fig:wafer}b) and c),
resulting in a total of $\SI{40000}{}$ images covering the whole wafer.

%-------------------------------------------------------------------------------
\subsection{Automated creation of an etch pit dictionary} 
%-------------------------------------------------------------------------------
In the following we introduce a data analysis pipeline for obtaining image 
regions that contain a single etch pit for a \gls{BPD}, a \gls{TSD}, and a \gls{TED}
with various Burgers vectors. 
In this part, the objective is \emph{not} to identify all etch pits in the 
wafer; the objective \emph{is} to automatically find a number of good examples 
(i.e., image sections with a single, clearly visible etch pit) that are suitable 
for creating semi-synthetic training data by superimposing such 
etch pit examples in an artificial image.

%-------------------------------------------------------------------------------
\subsubsection{Identification of image regions that contain an etch pit} 
\begin{figure*}
    \centering
    \includegraphics[height=0.9\textheight]{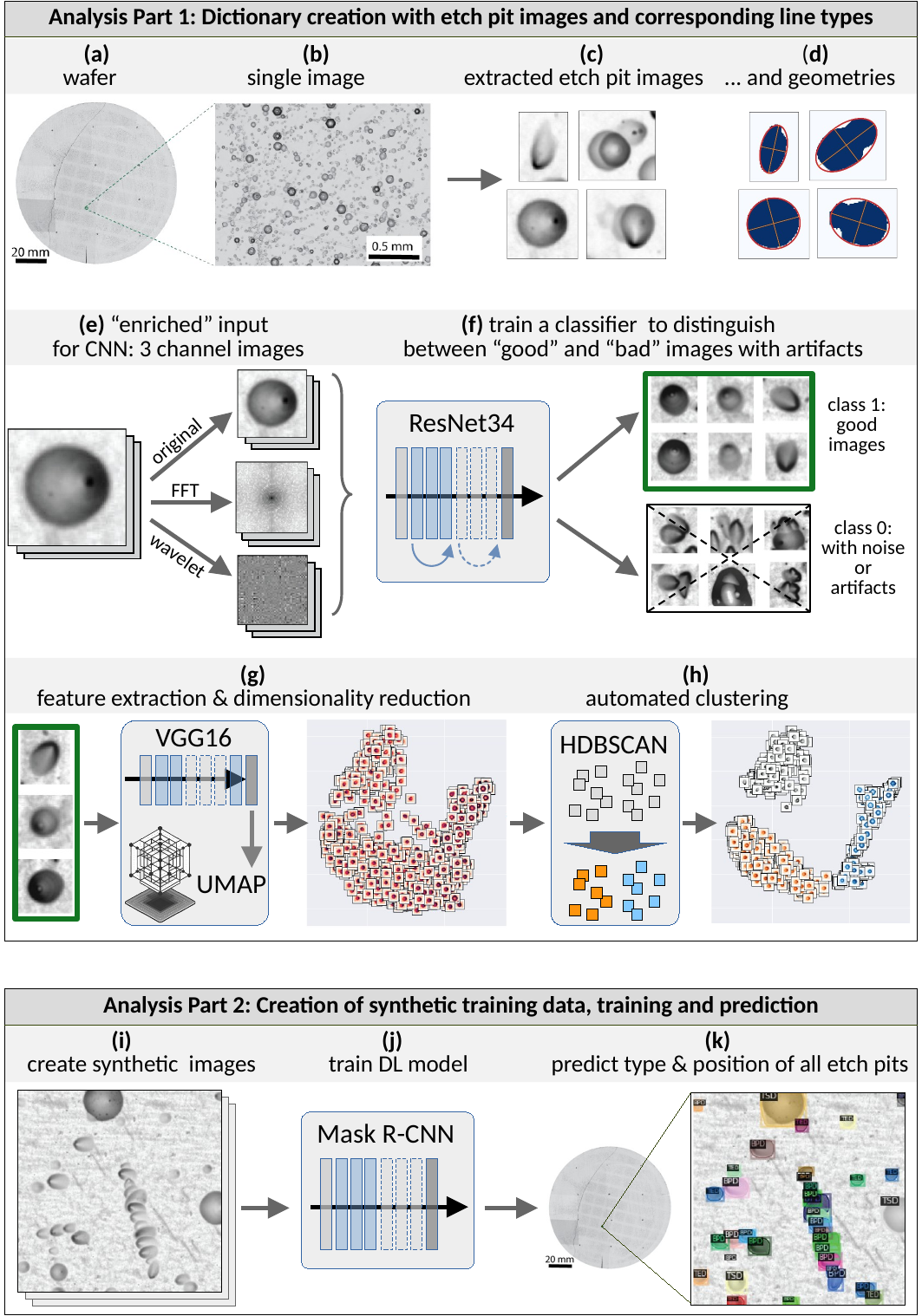}
    \caption[]{%
        Data analysis pipeline of the two main tasks: the automated creation of a etch-pit dictionary pool (top box) and predicting dislocations in the full wafer (bottom box). Further explanations are given in the text.
        % a) + b) real microscopy data from a full wafer; 
        % c) samples of extracted pixels group; 
        % d) three channels image data as input for training binary classification;
        % e: etch pits that are classified into $1$ and $0$; 
        % f: the $2$ components from $3$ reduced dimensions; 
        % g: grouping of etch-pits in different categories;
        % h: synthetic image data;
        % i: segmentation of three dislocation types.
    }
    \label{fig:schema_automate_clustering}
\end{figure*}

%-------------------------------------------------------------------------------
Each gray scale microscopy image (\cref{fig:schema_automate_clustering}b) 
consists of $1292\times968$ pixels. 
As a preprocessing step, distortions and inhomogeneous contrast were removed in each
image through the rolling-ball \citep{sternberg1983biomedical} and CLAHE 
(Contrast limited adaptive histogram equalization technique) method 
\citep{vidhya2017effectiveness} (see Appendix \ref{app:modified_image} for a brief explanation). 
A binarization threshold was then applied to the image to reveal the darker
etch pits (note, that during these steps it is not important that no all
etch pits are identified). Additional image processing techniques such 
as erosion, opening, and dilation using the OpenCV library were used for 
separating contiguous pixel groups (\cref{fig:schema_automate_clustering}c). 
% in the interest range of the grey scale values. 
To estimate the shape, a possibly rotated ellipse is fitted to each of the 
obtained pixel groups (\cref{fig:schema_automate_clustering}d). 
By characterizing the shape and size of the ellipse it is possible to exclude
pixel groups that do not correspond to etch pits. As shown in \citep{nguyen2023automated}
this requires the calculation of three parameters,
% the ``lengthiness'' of the ellipse ($c_1$), the ``compactness'' of 
% the pixel cluster ($c_2$) and the ``circularity'' ($c_3$). E.g., if the values of $c_1$, 
% to $c_3$ are greater equal $3$, $0.6$ and $0.6$, respectively, then the pixel 
% groups have a too extreme shape to be etch pits. 
the ``lengthiness'' of the ellipse, the ``compactness'' of 
the pixel cluster and the ``circularity''. If these characteristic values
exceed certain values (here: $3$, $0.6$, and $0.6$, respectively), then the pixel 
groups have a too extreme shape to be etch pits (see \citep{nguyen2023automated} for
further details).
These pixel groups act as a mask through which 
individual regions of interest containing a single etch pit are obtained (the 
masks were additionally expanded by a boarder of $10$ pixels). 
This process is done automatically for all \SI{40000}{} images covering the entire 
wafer and resulting in a total number of approximately \SI{1.7}{\text{million}} etch pit images. % $1,764,853$ images. 
However, not all of these etch pit images can be used for further analysis, 
because, e.g., two etch pits might overlap (class 0 in 
\cref{fig:schema_automate_clustering}f shows such examples), and therefore
it would not be possible to  uniquely determine the dislocation line character. 
To exclude these ``bad examples'' from the further analysis steps, a deep learning-based 
classification method is used, as introduced subsequently.
%Note, that it is not important to exactly detect all etch pits in this step. The objective here is to detect some of them and then to use them to create training images.
%

%-------------------------------------------------------------------------------
\subsubsection{Binary classification for selecting good candidates of etch pit images}
%-------------------------------------------------------------------------------
Generating synthetic training data by superimposing a number of each etch-pit images 
requires images of high-quality. Therefore, it is necessary to differentiate 
between a suitable and unsuitable candidate from the pool of extracted etch pits 
datasets obtained above. Suitable images contain only a single etch pit and no
artifacts, see the examples shown in \cref{fig:schema_automate_clustering}f.
%
%We propose a deep learning method for automating this task, the multiple channels convolutional neural network (CNN) with the backbone ResNet-34. In this work, we use three channels described in the Method section. 
%
To decide which of the images is a suitable image for further analysis, a 
\gls{CNN} is used as a classifier in a supervised learning setup. 

As network architecture a multiple-channel \gls{CNN} with a ResNet34 as a backbone 
was used. The input channels used for this type of \gls{CNN} consists of the 
original grey scale image, a magnitude spectrum of the Fast-Fourier transform of the 
image as well as of the wavelet transform (see \cref{fig:schema_automate_clustering}e). 
These three channels together contain 
significantly more information than just the gray scale values used in regular 
\glspl{CNN} and are very beneficial for the training and testing accuracy. 

As training dataset we select (by quick visual inspection which takes less than an hour) 
$1000$ arbitrary etch pit images that are clearly good candidates and $1000$ that are clearly unsuitable. The size of each image is $64\times64$ pixels. We also perform data augmentation such as rotating the image by various degrees as well as adding noise to the image to ensure that the model generalizes well. The resulting dataset is split into $5$ parts that are used for a $k$-fold cross-validation, where we ensured that there is no class imbalance. The test dataset is chosen as one of the five parts.
The performance of the trained network is evaluated in terms of classification accuracy, defined as the ratio between the sum of the correctly predicted records and the total number of predicted records. 

In the data analysis pipeline, the input to the trained network is one of the image
sections \cref{fig:schema_automate_clustering} c) and the output, i.e., the prediction of the network is the class label 0 or 
1. There, 0 indicates that the image section contains multiple overlapping etch 
pits or even other (image or crystallographic) defects while 1 indicates an 
image section that  contains exactly one etch pit and is suitable for further 
processing.

%-------------------------------------------------------------------------------
\subsubsection{Automated clustering of different dislocation types}
%-------------------------------------------------------------------------------
Subsequently, we continue with only those image section that were identified as 
good candidates for further analysis. The next goal is to identify the exact 
dislocation type (i.e., \gls{BPD}, \gls{TED} or \gls{TSD}), which is so far not known.
In principle this is again a classification task. However, labelling the training data (i.e., sorting all training images into several different dislocation categories) is very time consuming
time consuming and additionally prone to errors. 

A different approach consists in unsupervised learning where the training is not 
based on pre-assigned class labels. To sort etch pits images into various 
categories, clustering methods can help by automatically grouping similar images.
For determining what ``similar'' means, clustering methods often operate in a feature space which represents the images by a reduced set of essential features. 
Effectively, this implies an automated feature extraction followed by a dimensionality reduction, cf. \cref{fig:schema_automate_clustering}g.

For constructing this reduced feature space we firstly use VGG-16 neural network 
\citep{simonyan2014very} to convert etch pit images into feature vectors. No training
is done; instead, a pretrained network with weights taken from ImageNet-1k \citep{russakovsky2015imagenet} has been use for this purpose. VGG-16 is a 
simple convolutional neural network that consists of $13$ convolutional layers 
followed by three fully connected layers. The first fully connected layer is 
taken as the $4096$-dimensional feature vector and contains a 
multi-scale lower-level representation of the image which is suitable for 
clustering. However, the number of dimensions is still high and can cause problems for 
clustering methods due to the sparsity of the feature space. Thus, a further 
dimensionality reduction of the feature space is performed. 

% The ``work horse'' of unsupervised dimensionality reduction is \gls{PCA}
% \citep{jolliffe2002principal} which maps a $n$ dimension dataset to a new set of 
% $\leq n$ principal components. These are orthogonal and preserve the highest 
% possible variance of the data. This method is very efficient also for big 
% datasets, however, the principal components are not interpretable since the 
% algorithm is mainly driven by finding dissimilar samples 
% \citep{ringner2008principal}, and the global structure of the dataset is not preserved.
%
% \gls{t-SNE} \citep{vandermaaten08a} not only reduces the dimensionality of data but also preserves the original information. It keeps similar instances close while the dissimilar ones stay apart. However, it is a limitation when dealing with large datasets with many observations because of the computationally intensive.
%
%Since we want to deal with relatively large datasets, we need to find another method for this purpose with good performance. \gls{UMAP} is our choice.
%
A sophisticated method for unsupervised dimensionality reduction is 
the so-called \emph{\glsentrylong{UMAP}}\glsunset{UMAP} \gls{UMAP} \citep{mcinnes2018umap, becht2019dimensionality}. It preserves the global 
data structure and is still computationally manageable. The mathematical 
background of this technique is related to Laplacian eigenmaps 
\citep{belkin2003laplacian} which distribute data uniformly on (sub)manifolds of the data 
space. For more detail on the algorithm and description refer to 
\citep{mcinnes2018umap}.
%
%In our work, hyper-parameters are chosen to fit our purpose. 
As hyper-parameter of \gls{UMAP} we used 10 neighbours, 32 random states, and a minimum distance of $0.3$. The other parameters are taken as the default values of the library package \citep{mcinnes2018umap}. The $4096$-dimensional dataset from the VGG-16 feature vectors are then reduced to only three components (two of which are shown in \cref{fig:schema_automate_clustering}g). 

Finally, we use HDBSCAN (Hierarchical Density-Based Spatial Clustering of Applications with Noise) \citep{mcinnes2017accelerated} to automatically cluster the data of the three-dimensional feature space into three separate groups, which can then easily be identified as \gls{BPD}, \gls{TED} and \gls{TSD} as discussed below.
%(shown in grey, orange and blue colour, respectively). The result of the clustering can be seen in \cref{fig:schema_automate_clustering}h. 

%-------------------------------------------------------------------------------
\subsection{Predicting dislocations in all sections of the wafer}
%-------------------------------------------------------------------------------
At the end of the previous steps we have obtained a number of small images each
of which contains a single etch pit. We additionally know to which kind of dislocation
this etch pit corresponds. The images together with the types are contained in the ``etch-pit dictionary''. 
This allows now to generate synthetic image data based on these small images.

%-------------------------------------------------------------------------------
\subsubsection{Synthetic image generation}
%-------------------------------------------------------------------------------
Artificial images with etch pits are created by (i) creating a background and
(ii) randomly or systematically placing images from the dictionary on the background
and simultaneously creating a mask for the semantic segmentation.
%From the above methods, the single etch pit images for each of the dislocation types are automatically labelled and divided into the three categories \glspl{BPD}, \glspl{TED} and \glspl{TSD}. 

%\paragraph{Background generation}
Background images ($1024\times1024$ pixels) are grown by a non-parametric sampling method for texture synthesis \citep{efros1999texture, levina2006texture}. From an initial seed ($200\times200$ pixels), which is taken from a real microscopy image, the texture of the background image is grown one pixel at a time by a Markov random field model. Each newly synthesised pixel is generated from the centre of the chosen neighbourhood, which is similar to the pixel of the neighbourhood. These neighbourhoods are already found by the algorithm.

%\paragraph{Full image generation}
As a parameter, the minimum and maximum number of \glspl{BPD}, \glspl{TED}, and \glspl{TSD} that can appear in each of the images is defined. The algorithm randomly chooses the number of each dislocation type within these ranges. Each image is randomly selected from the image pool and pasted on the randomly chosen background image. An additional method is to
place dislocations along a straight line segment, mimicking low angle grain boundaries.
Altogether, around \SI{580000}{} etch pit images were used to create randomised training datasets.
The resulting synthetic images are obtained together with the mask for semantic
segmentation and with the position of each dislocation etch pit. We generate about \SI{10000}{} images for the training datasets and $1000$ images for validation. 
The size of each synthetic image is $512\times512$ pixels.

%-------------------------------------------------------------------------------
\subsubsection{Instance segmentation and classification of dislocation types}	
%-------------------------------------------------------------------------------
The instance segmentation and classification of dislocation types are done by training a Mask R-CNN \citep{he2017mask} deep learning model, which has been implemented in Detectron2 \citep{wu2019detectron2} (a Facebook AI Research's next-generation library that provides detection and segmentation algorithms), which is built based on the Pytorch platform.
Mask R-CNN \citep{he2017mask} extends Faster R-CNN \citep{ren2015faster} by adding a branch for predicting segmentation masks on each Region of Interest (RoI) in parallel with the existing branch for classification and bounding box regression. The mask branch is a small fully connected network applied to each RoI, predicting a segmentation mask in a pixel-to-pixel manner. Mask R-CNN is simple to implement and to train given the Faster R-CNN framework, which facilitates a wide range of flexible architecture designs. Additionally, the mask branch only adds a small computational overhead, enabling a fast experimentation.

In our implementation we train the model with a ResNet101 as backbone using pre-trained COCO weights. 
The dataset is converted into COCO format, in which the annotations contain a list of dictionaries with the required information, suitable for Detectron2.
The performance of the segmentation is evaluated in terms of the \gls{RMSE}, which is
defined as
\begin{align}
    \label{eq:rmse}
    e = \sqrt{\frac{1}{N} \sum_{i=1}^{N} (y_{i}-\hat{y}_{i})^2}\;,
\end{align}
where $N$ is the number of values, $y_{i}$ is the $i$-th truth value and $\hat{y}_{i}$ is the $i$-th predicted value.
For the evaluation of the performance, $5$-fold cross-validation was conducted 
giving for each fold the accuracy of $0.89$, $0.92$, $0.96$, $0.94$ and $0.92$ 
respectively, which results in an average accuracy of $\approx 0.92$.

This was the last step in this lengthy data analysis pipeline. We are now able to
detect the position and dislocation type in each of the images of the full
wafer, as shown in \cref{fig:schema_automate_clustering}k. The Burgers vector
of the respective defect can be obtained in a conventional manner by computing
the radius of the etch pit (compare \cite{nguyen2023automated}).

%===============================================================================
\section{Results and discussion}
%===============================================================================
In the following we start by investigating the physical soundness of the 
data analysis results. Afterwards, the dislocation predictions are discussed 
within the materials scientific context.

%-------------------------------------------------------------------------------
\subsection{Robustness of the clustering}
%-------------------------------------------------------------------------------
%
\begin{figure*}[htp]
    \centering
    \includegraphics[width=1\textwidth]{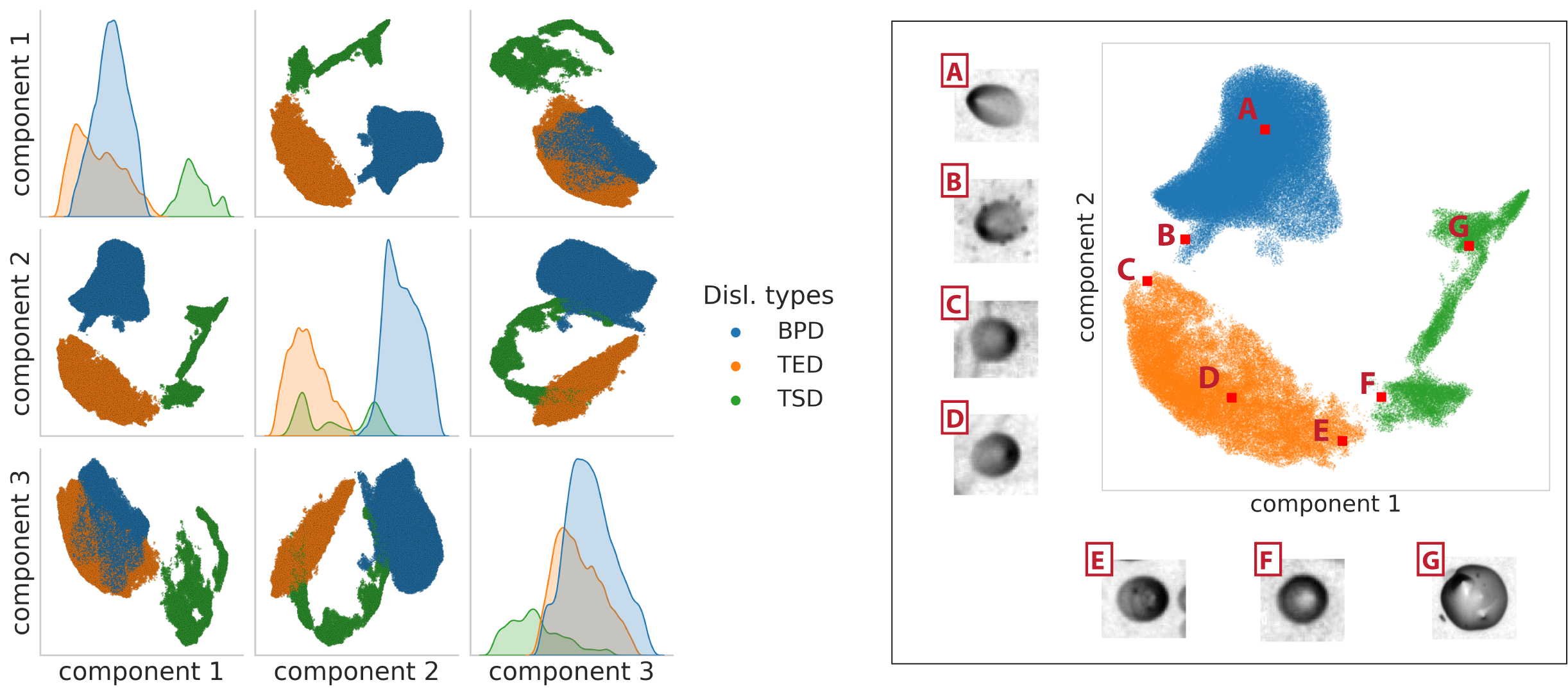}
    \caption[]{Left: Visualization of all three components after clustering.
    Right: Magnification of the data projected on the first two components (i.e., the middle plot in the leftmost column of the left figure) along
    with examples of the etch pit images. The value ranges of the three latent space 
    components are the same, and therefore, no scales are given.}
    \label{fig:component_and_count_label_dist}
\end{figure*}
The first part of above introduced data analysis pipeline (etch pit dictionary creation) was designed to be as robust as possible.
In particular, it was designed such that missing even a larger fraction of
etch pits should not have any impact on the quality of the dictionary. 
As an important part we will now investigate the clustering results,
\cref{fig:schema_automate_clustering}h in more detail.

There are three different dislocation types, which commonly appear in microscopy images of SiC after etching with molten KOH:  \glspl{BPD} have a sea-shell-like shape, while \gls{TED} and \gls{TSD} have a round or hexagonal shape with a core in the center region. Larger cross sections indicate larger Burgers vectors.
%
%Each etch pit image was encoded into a $4096$ feature vector obtained from the VGG-16 Net. This vector was then mapped to a three-dimensional manifold using \gls{UMAP}, resulting in a latent space that is then suitable for the clustering process.
%
In \cref{fig:schema_automate_clustering}g and the magnification in Appendix~\ref{app:magnification_image} we observe that similar etch pits are located close to each other in the latent space generated by \gls{UMAP}, which is beneficial for the clustering process.
\Cref{fig:schema_automate_clustering}h shows three clusters which are obtained by applying the HDBSCAN clustering method to the three-dimensional \gls{UMAP} latent space. For 
visualization purposes, they are further reduced to two dimensions, shown in grey, orange 
and blue colour respectively. Note, that (again for visualization purposes) in 
\cref{fig:schema_automate_clustering}g we only show $1000$ data points from part $9$ of the 
wafer with embedded images of those etch pits.

%---------------------------------------------------------------
How accurate was the clustering? Do the different colors in fact correspond to the
three types of dislocation (\glspl{BPD}, \glspl{TED} and \glspl{TSD})?
To answer this question, we take a look at the correlation between the three feature components shown in \cref{fig:component_and_count_label_dist}.
%with the labels corresponding to three types of dislocation (\glspl{BPD}, \glspl{TED} and \glspl{TSD}). 
%They are clustered into groups $0$, $1$ and $2$, respectively. 
%
First of all, we observe that three components are sufficient to distinctly
separate the point clusters: E.g., in the plot of component 1 vs. 3 \glspl{TSD} 
are even linearly separable from the other dislocation types. Similarly, 
plotting components 1 vs 2 and ignoring the \glspl{TSD} makes \glspl{BPD} and \glspl{TED}
linearly separable as well. 

Exploring the structure of the latent space can also be done by taking a look at 
how the corresponding images change when moving from one cluster to the next.
\Cref{fig:component_and_count_label_dist} (right) shows several example images in the latent space. Position A and G both indicate very extreme shapes: both are rather large and have a strong contrast. However, A belongs to the group of lengthy shapes (the basal plane dislocations) while G belongs to the threading screw dislocations and is rather round.
In between is the group of threading edge dislocations. They are smaller than the \glspl{BPD} but still roundish. The clustering method successfully separates even those
images that are close to each other, e.g., E and F. Visual inspection shows that E and F are indeed different. The same holds for B and C which implies that the three groups of dislocations can be unambiguously separated, making the method very robust for this application.

\subsection{Reliability of predicting various dislocation types}
%-------------------------------------------------------------------------------
\begin{figure*}[htp]
    \centering
    \includegraphics[width=1\textwidth]{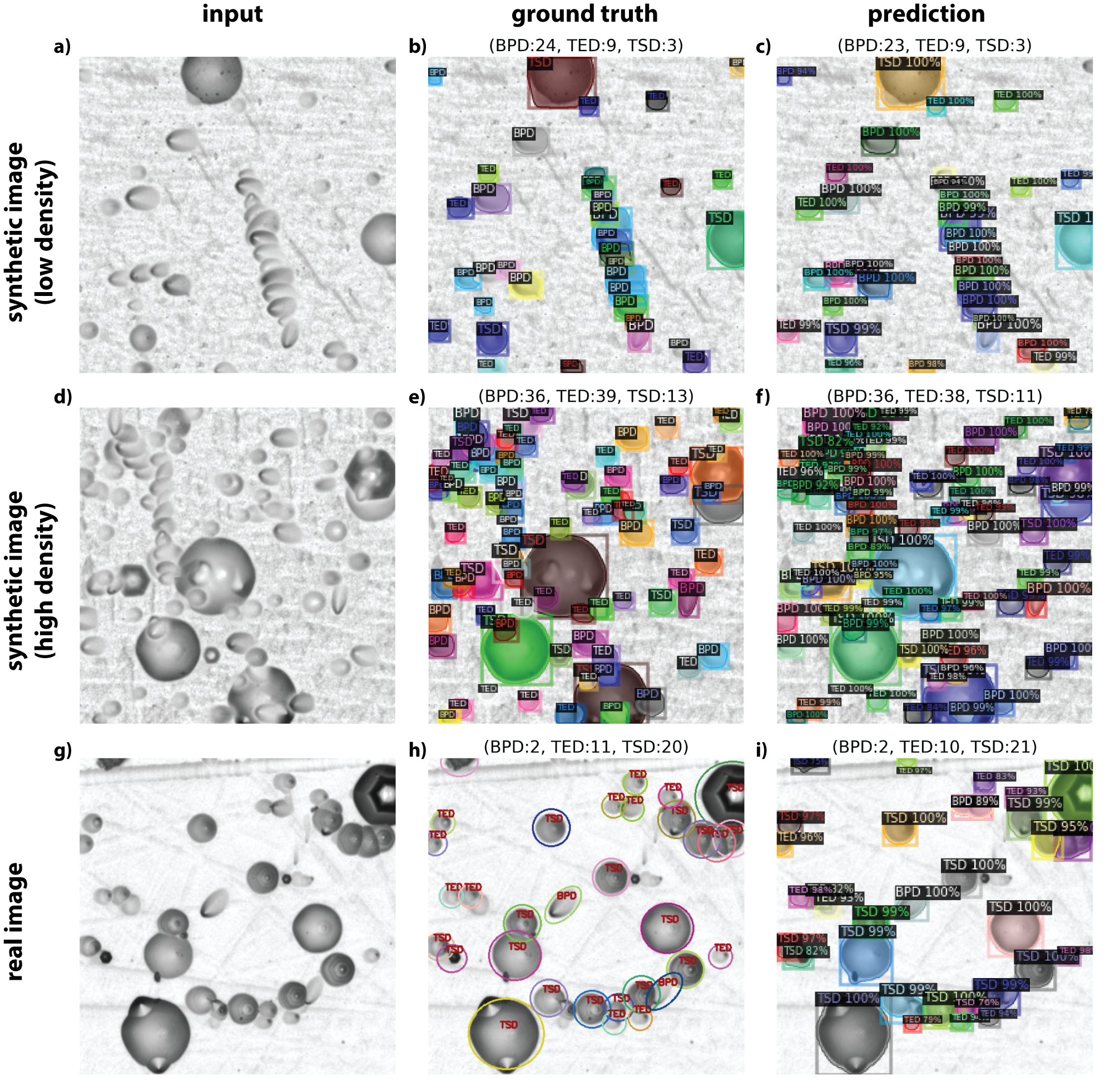}
    \caption[]{Results comparison between ground truth and predicted segmentation. a and d: grayscale images of synthetic low and high dislocation density; g: real microscopy image; b and e: ground truth segmentation from the synthetic image; h: ground truth segmentation from hand labelling; c, f and i: predicted results from the deep learning. }
    \label{fig:predict_on_test_image}
\end{figure*}
For the purpose of validating the trained segmentation model three different 
datasets were created that were not used in the training process: 
\begin{itemize}
    \item The first dataset contains $1000$ synthetic 
    images with a low etch pit density where the number of etch pits were randomly
    varied within the ranges of $0..5$ \glspl{TSD}, $0..10$ \glspl{TED}, and 
    $0..20$ \glspl{BPD}. 
    
    \item The second dataset again consists of $1000$ synthetic images. However, 
    each of them contain, on average, a larger number of etch pits. Here, we used the 
    ranges of $0..20$, $0..50$ and $0..200$ for the quantities of the three dislocation types. 
    
    \item The third dataset consists of $100$ real microscopy images that 
    were manually annotated by a domain expert.
\end{itemize}
\Cref{fig:predict_on_test_image}a and \ref{fig:predict_on_test_image}d 
show examples of synthetic microscopy images for the case of low 
dislocation density and high dislocation density, respectively. The images 
contain all three types of dislocations, for which a number of etch-pit images are randomly 
chosen from the dictionary. The dislocation images of high density are 
unrealistic since such densities are typically not found in commercial SiC 
wafers. However, including this extreme case turned out to be beneficial for 
the training process of 
the instance segmentation since the difficult cases increase the variance of 
the dataset and thereby helps the model to generalize better.

\begin{figure*}
    \centering
    \includegraphics[width=1.0\textwidth]{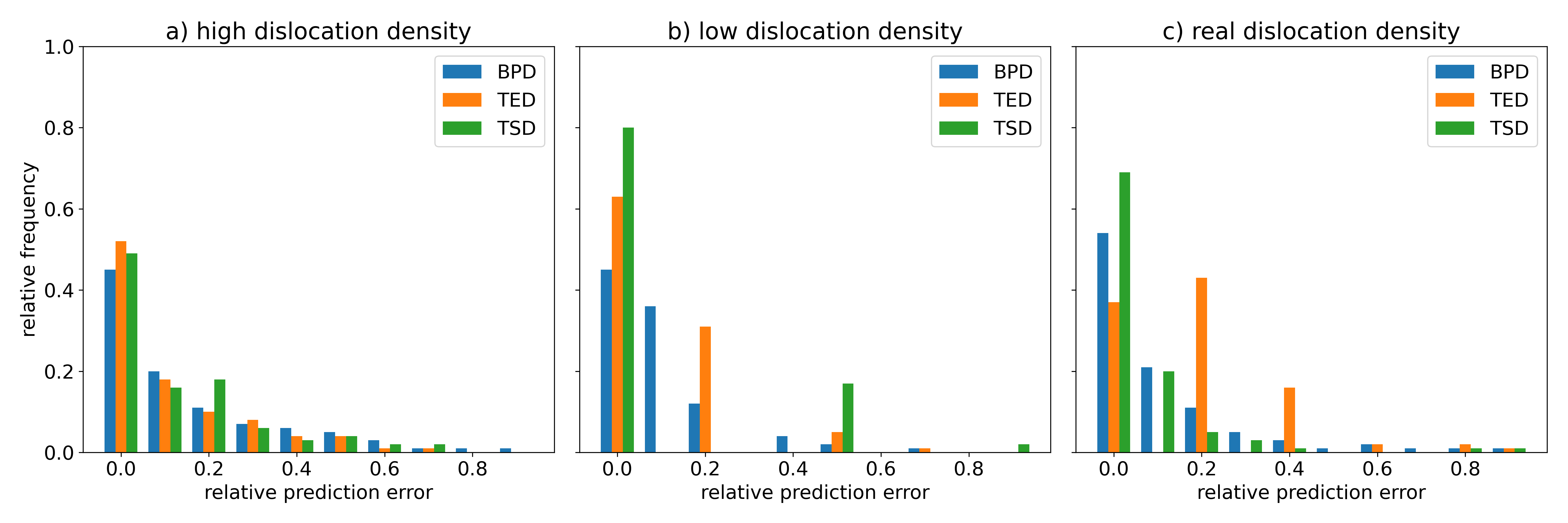}
    \caption[]{Differences between ground truth and prediction on various test datasets: $1000$ images of high dislocation density, $1000$ images of low dislocation density and $100$ images of real dislocation density.}
    \label{fig:plot_differences}
\end{figure*}

The  ground truths and predictions are shown in the middle and right column of \cref{fig:predict_on_test_image}.
To quantify the error the  \gls{RMSE} %(cf. \cref{eq:rmse}) 
is calculated for 
all dislocation types resulting
in $e_\text{BPD}=13$, $e_\text{TED}=23$, and 
$e_\text{TSD}=3$ for the case of high dislocation density. 
These values are much smaller for the case of lower dislocation density which 
are $e_\text{BPD}=1$, $e_\text{TED}=4$, and 
$e_\text{TSD}=0.5$, respectively.
For the real images, the \gls{RMSE} values are obtained as
in $e_\text{BPD}=3$, $e_\text{TED}=5$, and 
$e_\text{TSD}=2$.
Analyzing the prediction performance of the model for all images of the validation
datasets results in the distributions of absolute classification errors as 
shown in \cref{fig:plot_differences}.
The dataset with the high dislocation results \emph{by design} in the worst 
performance: the dataset was created to be the most challenging one. 
By contrast, the validation dataset with the lower density the model shows a very 
good prediction accuracy. The ultimate test, however, is the real dataset
since it is a subset of the whole wafer. For this, the variance of the errors is
larger than that for the low density synthetic dataset. However, on average,
only $1.9$ \glspl{BPD}, $0.9$ \glspl{TED}, and $1.3$ \glspl{TSD} were misclassified.
In many cases, there even was no misclassification at all. Assuming that this dataset 
is statistically representative of the full waver it can be assumed that 
this high level of accuracy also will hold for the
analysis of the full wafer.

%-------------------------------------------------------------------------------
\subsection{Analysis of the full wafer}
%-------------------------------------------------------------------------------
\begin{figure*}
    \centering
    \includegraphics[width=1\textwidth]{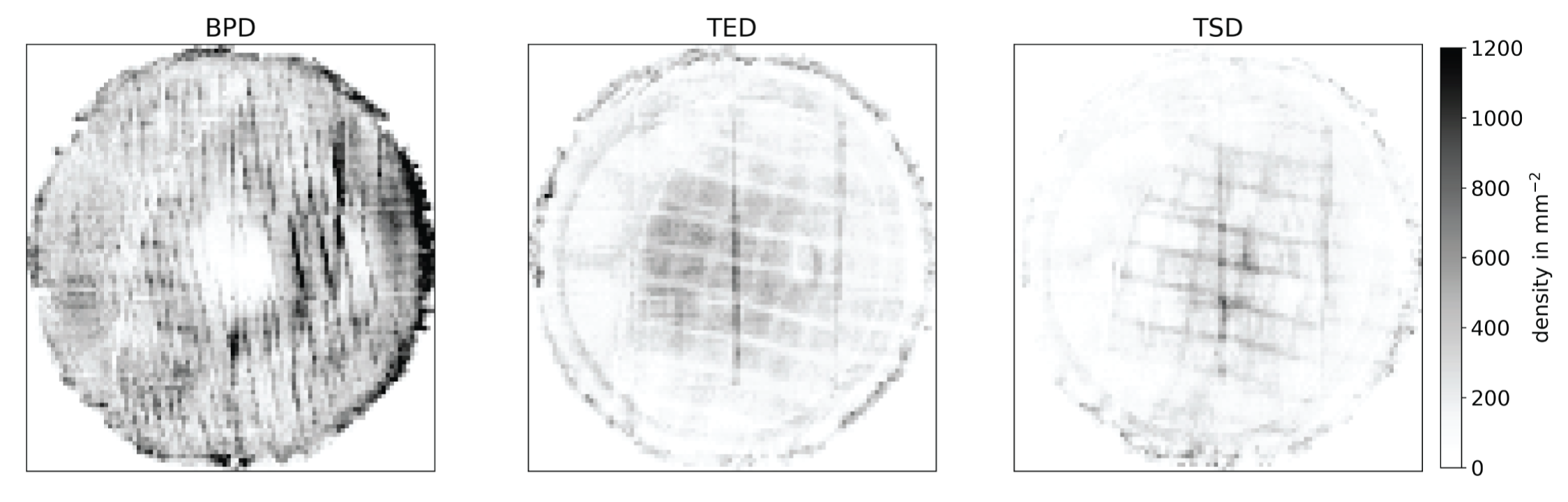}
    \caption[]{Visualization of the three types of dislocation density in the whole wafer.}
    \label{fig:hist2D_plot_disl_types_grey}
\end{figure*}
Analyzing the full waver is a formidable task since it consists of altogether
about \SI{40000}{} individual images. The above introduced data analysis pipeline 
is suitable for such a task because even though training of the Mask R-CNN network
is computationally expensive, making predictions about location and type of 
edge pits in new images is significantly faster.

Altogether $1.7$ millions etch pits were identified and located. 
\Cref{fig:cat_plot_count_part_index_vertical} in the appendix summarizes the dislocation content
for each of the 20 wafer subdivision, each of which again consists of a large
number of individual images. \Glspl{BPD} are the predominant type of
dislocation in most of the regions while on average \glspl{TED} occur 50\% less.
The number of \glspl{TSD} is the lowest in all wafer parts.

To 
illustrate the result of this analysis the resulting dislocation density 
distribution of the three dislocation types obtained for the whole wafer is 
shown in \cref{fig:hist2D_plot_disl_types_grey}. 
The density distribution of \gls{BPD} is the highest with a maximum value of around 
\SI{1.6e5}{\cm^{-2}}, while the maxima of \gls{TED} and \gls{TSD} are approximately 
\SI{1.2e5}{\cm^{-2}} and \SI{0.8e5}{\cm^{-2}}, respectively. 
We observe that \glspl{TED} and \glspl{TSD} are mainly concentrated in the central region and on the boundary of the wafer. The center of the wafer is mainly \glspl{BPD}-free and then takes a high value.
Due to the overlapping of the image sections, artifacts from imperfect image registration 
in form of two vertical lines occur (cf. middle and right figure of \cref{fig:hist2D_plot_disl_types_grey}).
The slightly tilted vertical lines in the left plot are arrays of \glspl{BPD}, 
forming several low angle grain boundaries that are distributed approximately in an equidistant manner. They occurred due to the high curvature of the growth interface and the shear stress acting on the grown crystal \citep{wellmann2023processing, nakano2019formation, steiner2023prevention} and became visible only through
this analysis. Furthermore, a varying TED density can be seen on the upper left side approximately every $30^0$ originating from the center of the wafer.

%-------------------------------------------------------------------------------
\subsection{Are there any dislocations missing?}
%-------------------------------------------------------------------------------
In our study, we assumed that most of the etch pits are either \glspl{BPD}, \glspl{TED} or \glspl{TSD}. However, in reality, there can also be dislocations of mixed line characters. This type of etch-pit occurs because of the switching direction of dislocations. E.g. a \gls{TED} can be deflected into a \gls{BPD}, or vice versa, due to  macro-step formation \citep{li2022investigation}. Similarly, \glspl{TSD} can bent into \glspl{BPD} and vica versa (``macro step flow'') \citep{mitani2021massive}. 
Furthermore, ``shallow dislocations'' were observed to form a round pit (with a curved bottom) without a core. Since dislocations cannot start or end inside the crystal, this structure can result from a dislocation half-loop  \citep{ishikawa2011detection}. For those pits, all dislocations are somewhat tilted \citep{tsubouchi2016characterizations}.

Clearly, all of these cases are not explicitly considered in our analysis and are
rather assigned to one of the three classes. Here, the clustering results 
might help to identify particular etch pit shapes by, e.g., taking a look at the
extreme ends of a cluster, similar to what was done in \cref{fig:component_and_count_label_dist}. An other option to extent this
investigation is to further cluster the content of a single cluster of images and thereby
to obtain further subdivisions of that cluster. This might be a good
starting point to detect new sub-groups of etch pit images that are similar in some aspects, which could be, e.g., the case for a mixed dislocation.

%===============================================================================
\section{Conclusion}
%===============================================================================
We have proposed a complex image analysis pipeline that consists of a number
of classical image analysis techniques, supervised deep learning methods as well
as unsupervised approaches. A particular emphasize was the robustness and accuracy of the framework. Generation of semi-synthetic training data was key for training an instance segmentation neural network without manually creating data annotation. 
This helped to rapidly analyze thousands of images and obtain accurate and
spatially resolved information about the distribution of different dislocation types. 

Our method can be applied to quickly identify areas of defect density inhomogeneities, for example, areas of high \gls{BPD} density due to localized stress during crystal growth. 
The increasing diameter standard of SiC wafers of up to \SI{200}{\mm} 
directly implies that KOH etching images have to be analyzed automatically 
since such a high amount of data can not be handled otherwise. 

Ultimately, high-throughput analysis methods as the one in this work 
will contribute to  understanding of defects in SiC which is the basis for optimizing the process parameters during growth.

	%\appendix

	
	%\bibliographystyle{cas-model2-names}
	
	%\bibliography{literature}

	\backmatter
	
	%\bmhead{Supplementary information}
	%
	%If your article has accompanying supplementary file/s please state so here. 
	%
	%Authors reporting data from electrophoretic gels and blots should supply the full unprocessed scans for key as part of their Supplementary information. This may be requested by the editorial team/s if it is missing.
	%
	%Please refer to Journal-level guidance for any specific requirements.
	
	%\bmhead{Acknowledgments}
	%Financial support of the Deutsche Forschungsgemeinschaft (DFG) under the contract numbers DA357/7-1, WE2107/15 and SA2292-6 is greatly acknowledged.
	%%
	%%Please refer to Journal-level guidance for any specific requirements.
	%%
	%\bmhead{Conflict of interest}
	%On behalf of all authors, the corresponding author states that there is no conflict of interest.
	
	\section*{Acknowledgments}
	Financial support of the Deutsche Forschungsgemeinschaft (DFG) under the contract numbers DA357/7-1, WE2107/15 and SA2292-6 is greatly acknowledged.
	
	\section*{Conflict of interest}
	On behalf of all authors, the corresponding author states that there is no conflict of interest.
	
	\section*{Data availability}
	Postprocessed data is available at zenodo, raw microscopy data is available upon request.
	
	\section*{Code availability}
	Code is available upon request.
	
	%\section*{Declarations}
	%
	%Some journals require declarations to be submitted in a standardised format. Please check the Instructions for Authors of the journal to which you are submitting to see if you need to complete this section. If yes, your manuscript must contain the following sections under the heading `Declarations':
	%
	%\begin{itemize}
	%\item Funding: The financial support of this work is from the Deutsche Forschungsgemeinschaft (DFG) under the contract numbers DA357/7-1, WE2107/15 and SA2292-6.
	%\item Conflict of interest: On behalf of all authors, the corresponding author states that there is no conflict of interest.
	%%\item Ethics approval 
	%%\item Consent to participate
	%%\item Consent for publication
	%\item Availability of data and materials: Data available upon request.
	%\item Code availability : Code available upon request.
	%%\item Authors' contributions
	%\end{itemize}
	%
	%\noindent
	%If any of the sections are not relevant to your manuscript, please include the heading and write `Not applicable' for that section. 
	
	%\bigskip
	%\bigskip\noindent
	%Springer journals and proceedings: \url{https://www.springer.com/gp/editorial-policies}
	%
	%\bigskip\noindent
	%Nature Portfolio journals: \url{https://www.nature.com/nature-research/editorial-policies}
	%
	%\bigskip\noindent
	%\textit{Scientific Reports}: \url{https://www.nature.com/srep/journal-policies/editorial-policies}
	%
	%\bigskip\noindent
	%BMC journals: \url{https://www.biomedcentral.com/getpublished/editorial-policies}
	%\end{flushleft}
	%

    \onecolumn

	\begin{appendices}
%===============================================================================
\section{Classical image analysis: Modifying the distortion and the unequal contrast} 
\label{app:modified_image}
\begin{figure}[!htb]
    \centering
    \includegraphics[width=0.5\textwidth]{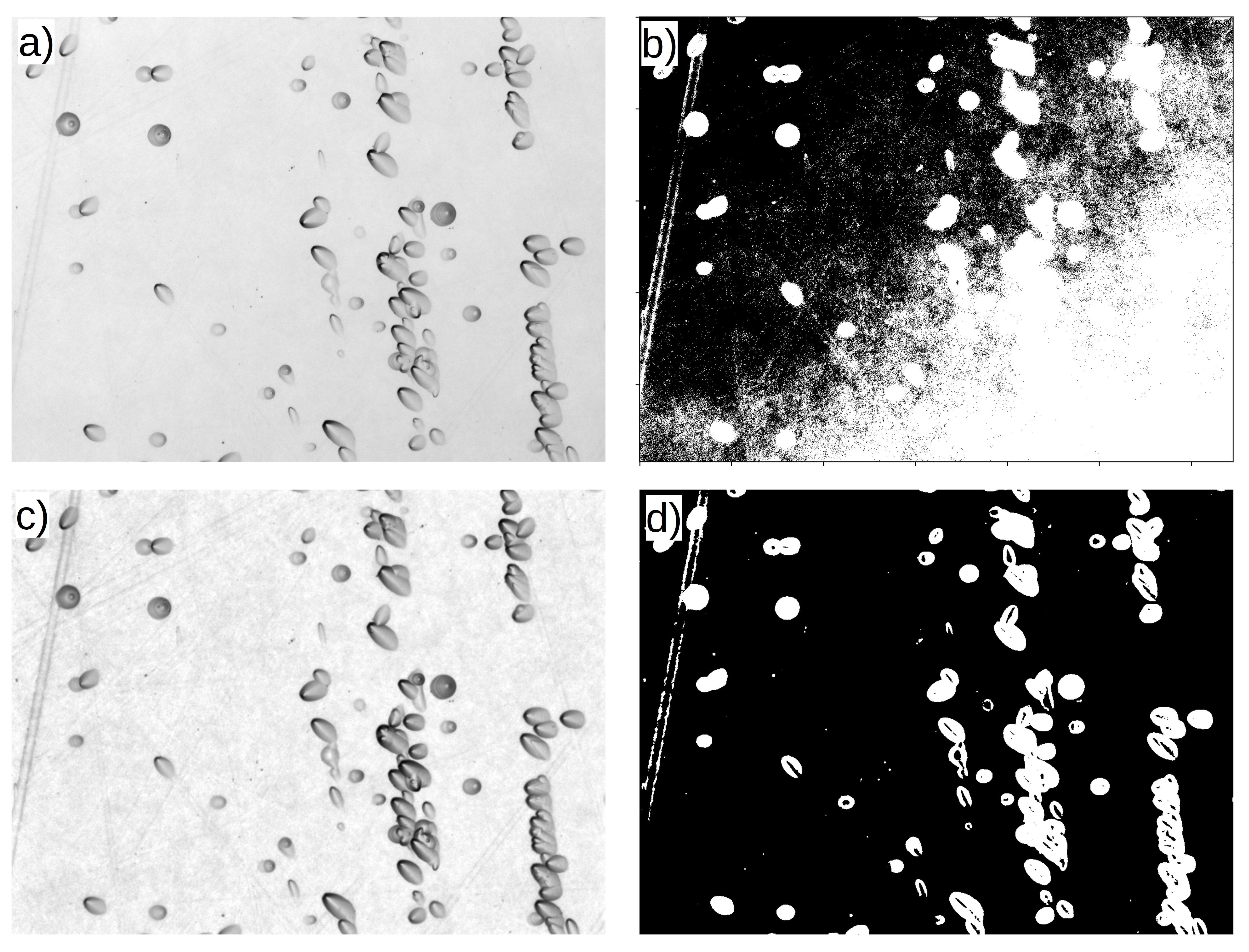}
    \caption[]{Correction of unequal contrast of the microscopy image: a) the original image; b) threshold of original image; c) the modified image; d) threshold of modified image.}
    \label{fig:shade_corrected}
\end{figure}

%===============================================================================
In a first step we enhance the quality of the original images. By applying the rolling-ball \citep{sternberg1983biomedical} and CLAHE (Contrast limited adaptive histogram equalization technique) \citep{vidhya2017effectiveness}, the features (etch-pit) are becoming more clear and the distortion which causes an unbalance to the contrast of the image are removed. The results are shown in \cref{fig:shade_corrected}. 
We can see that the unbalance of the contrast in the original image (see \cref{fig:shade_corrected}a) causes a non-optimal result when we take threshold this image. A large number of features of the etch pits in the lower right corner are lost (see \cref{fig:shade_corrected}b). On the other hand, when we perform the modification, the shading of the image looks more equal across the image (see \cref{fig:shade_corrected}c), with the result of detailed features of dislocation etch pits in the whole image when applying the threshold (see \cref{fig:shade_corrected}d).
		
%	\section{Binary classification with three channels image}
%		\Cref{fig:binary_classification} shows a sketch of the classification process with an example of the three channels image from an etch-pit, which consists of the original gray-scale image, the Fourier and the wavelet transformation of the original image. More composed details of this image type can refer to the work of \citet{nguyen2023efficient}.	
%		\begin{figure}[!htb]
%			\centering
%			\includegraphics[width=1\textwidth]{figures/binary_classification_3_channels.png}
%			\caption[]{Overview of binary classification with three channels image.}
%			\label{fig:binary_classification}
%		\end{figure}

%===============================================================================
\section{Detailed images from the dimensional reduction and automated  clustering}
\label{app:magnification_image}	
%===============================================================================

\begin{figure*}
    \centering
    \includegraphics[height=0.43\textheight]{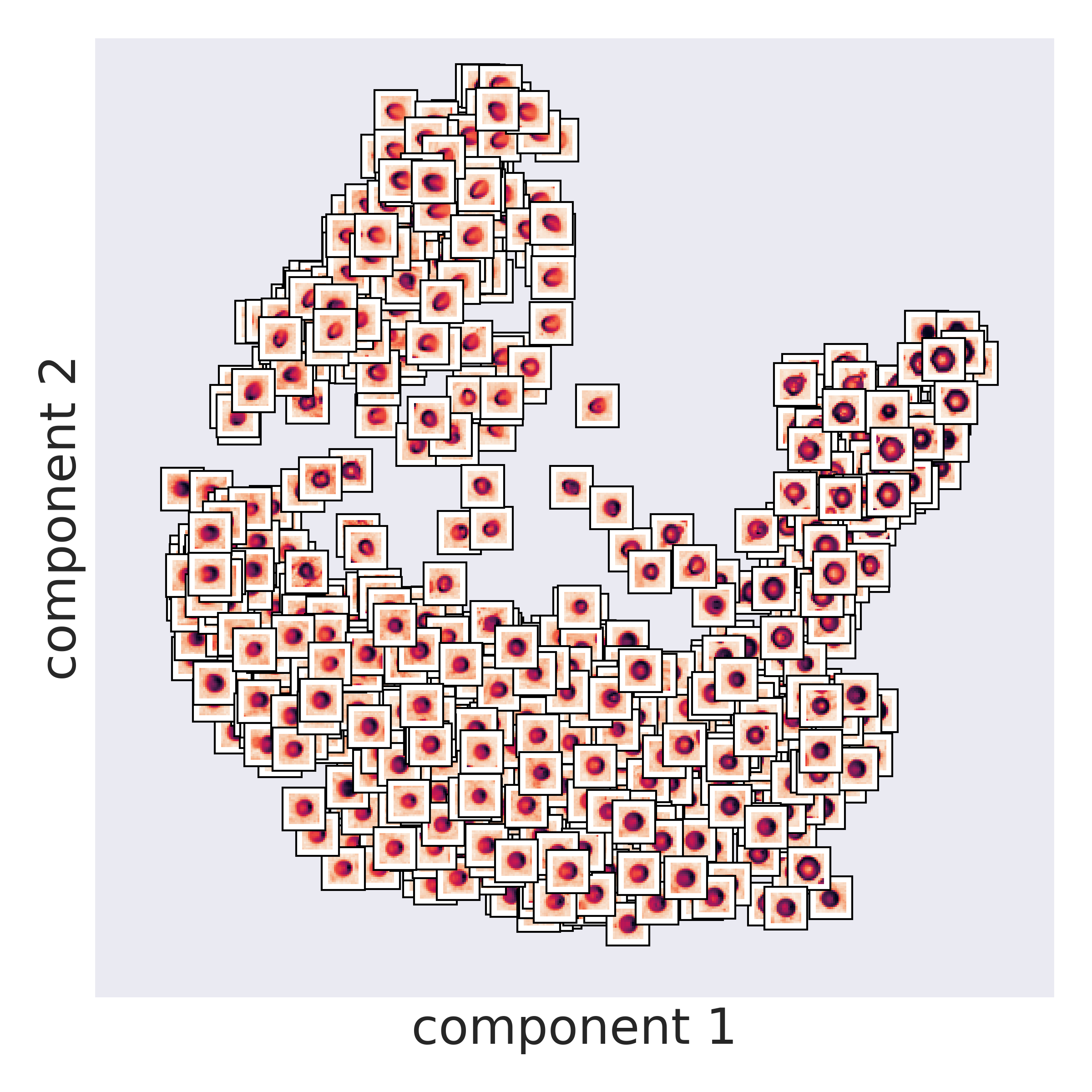}\\
    \includegraphics[height=0.43\textheight]{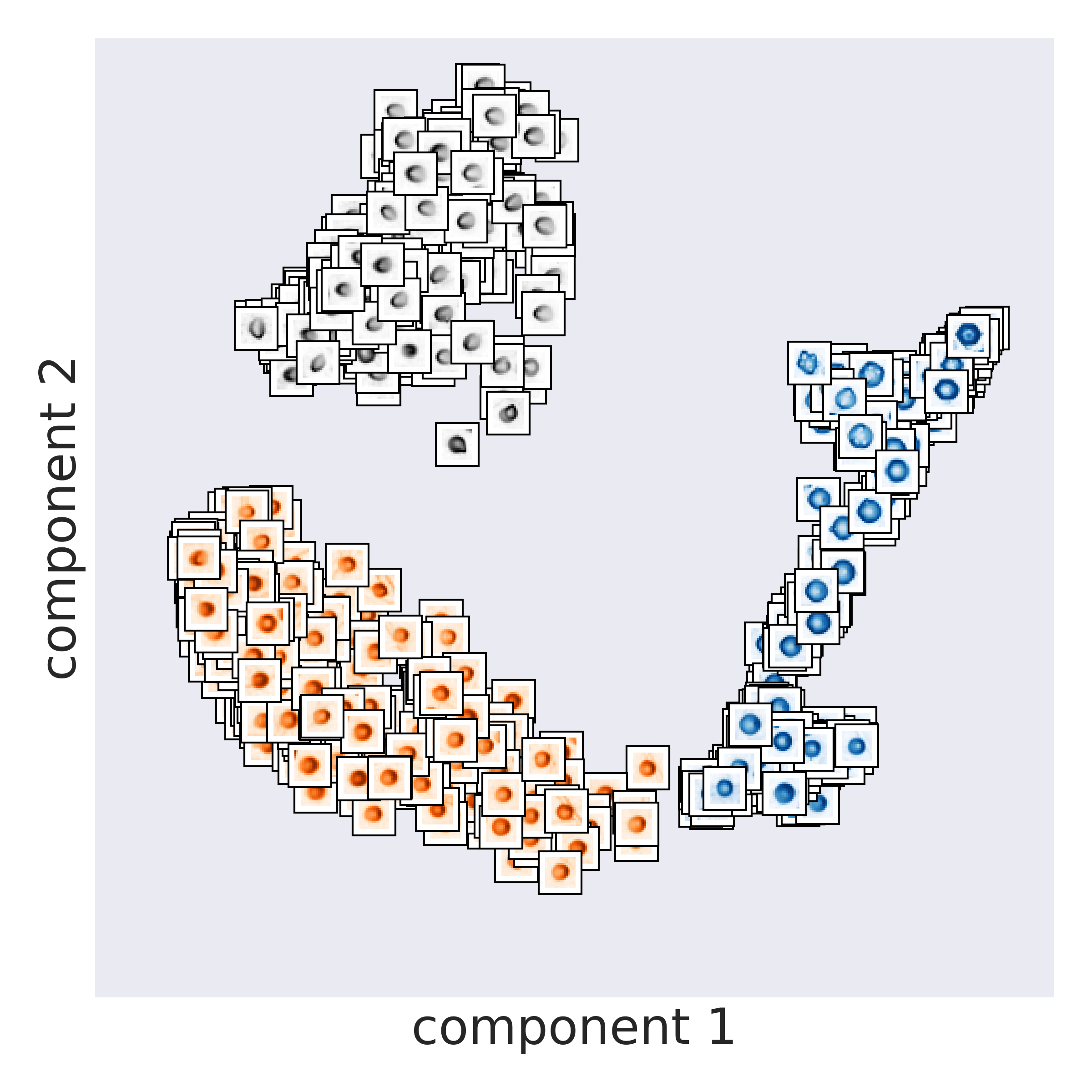}
    \caption[]{
        Visualization of the etch pit distribution after reducing the 
        dimensionality to three latent variables (the top figure shows the first 
        two variables) and after clustering (bottom). It can be seen that already 
        the dimensionality reduction by UMAP significantly sorts the images, e.g., 
        the most circular etch pits are located at the right. After automated 
        clustering, the three different groups of images are clearly separated.
    }
    \label{fig:automated_clustering}
\end{figure*}

\begin{figure*}
    \centering
    \includegraphics[width=0.7\textwidth]{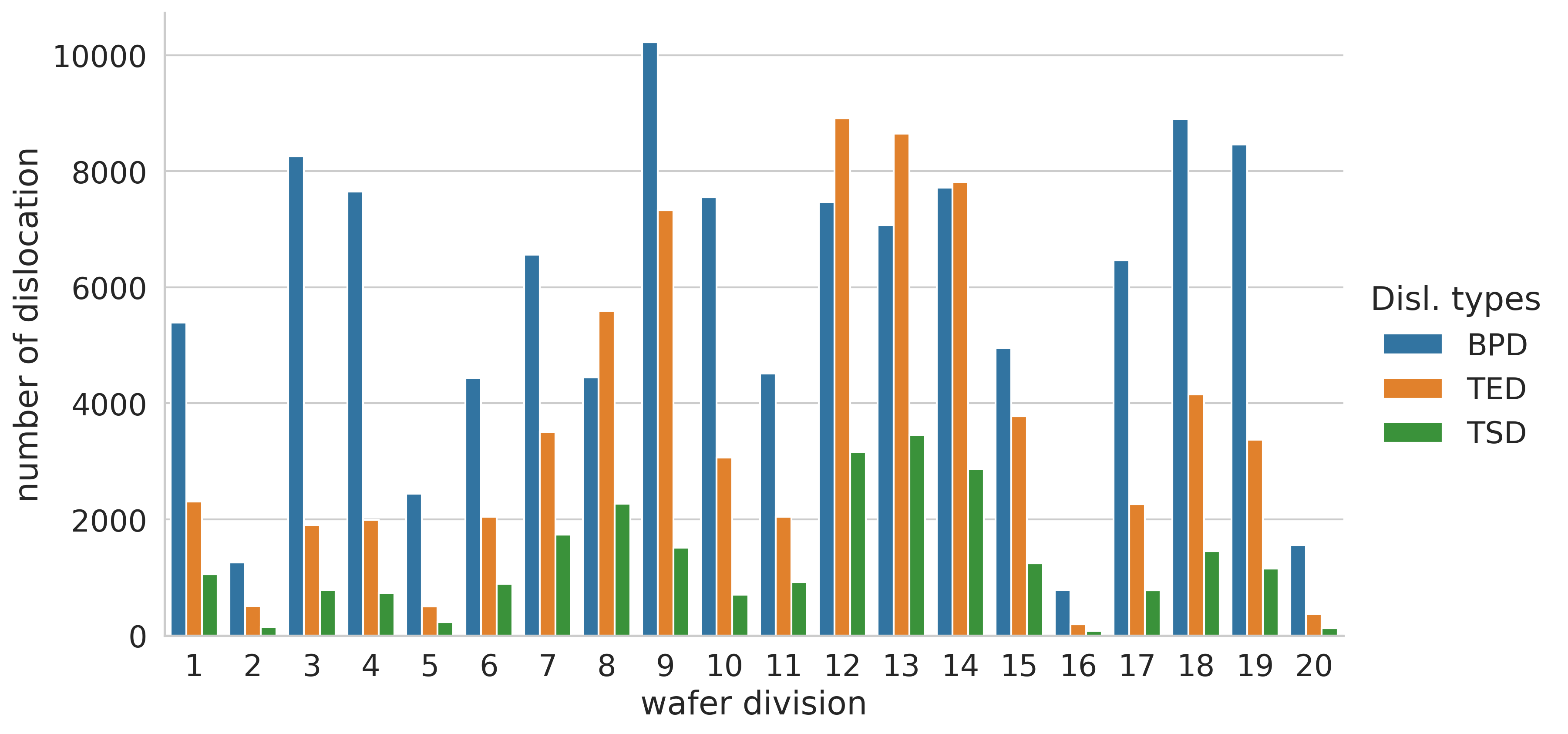}
    \caption[]{Counting etch-pits distribution on each wafer division.}
    \label{fig:cat_plot_count_part_index_vertical}
\end{figure*}

\Cref{fig:cat_plot_count_part_index_vertical} summarizes the dislocation content
for each of the 20 wafer subdivision, each of which again consists of a large
number of individual images. Obviously, BPD's are the predominant type of
dislocation in almost all regions (exception: part 8, 12 and 13, see
\cref{fig:wafer} for the location of these parts).

%\subsection{left over}
%Segmenting, classifying and annotating $100$ real images with the size $512\times512$ pixels would take a couple of hours for a human to do the jobs while it only takes $14$ second for our trained code to predict all information from the image, such as types number, type, size, position of etch pits that appear in the images. 

%===============================================================================
\section{Unsupervised clustering results from various combination of Deep Learning techniques and clustering approaches}
\begin{figure*}[!htb]
			\centering
			\includegraphics[width=1\textwidth]{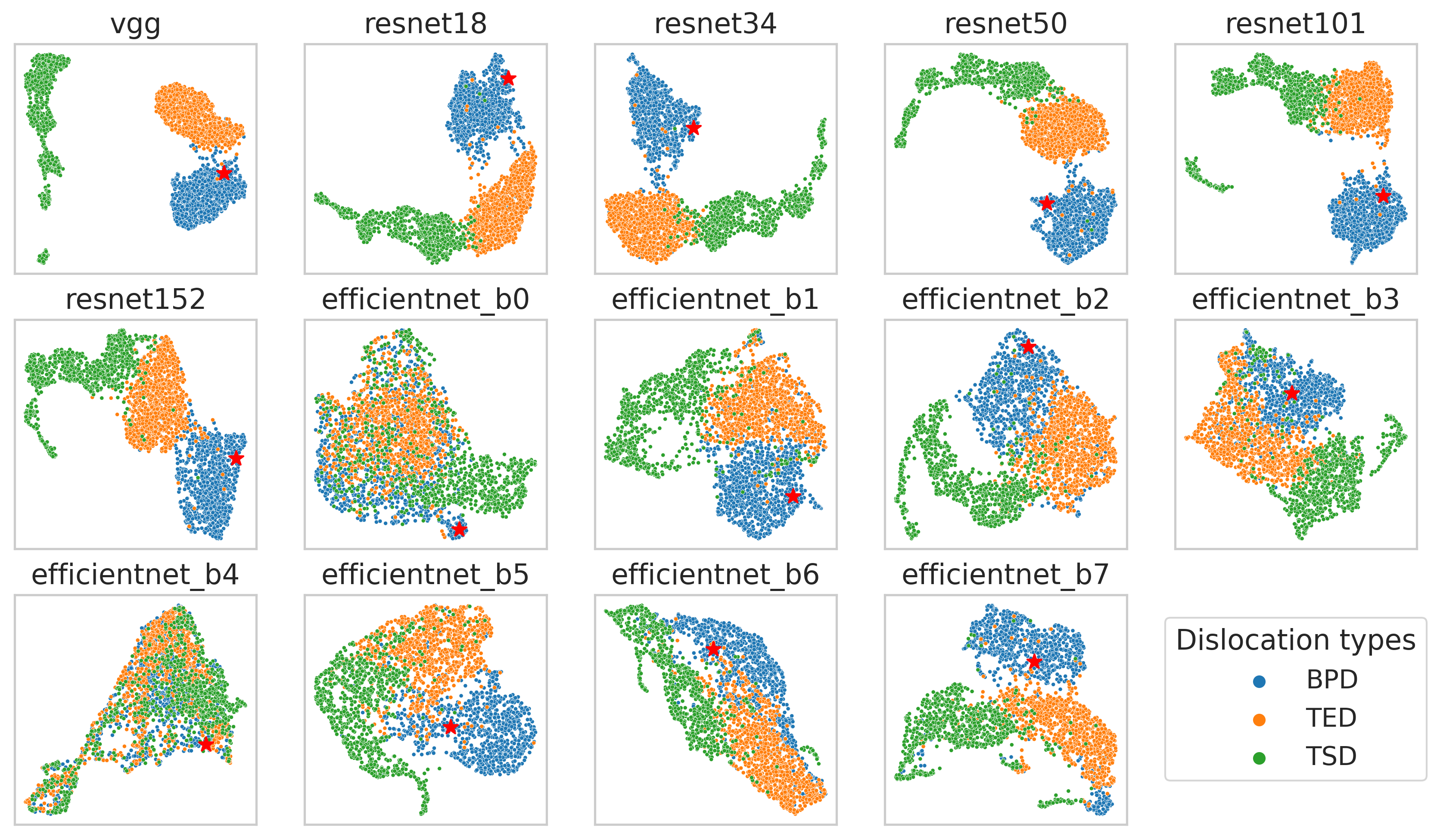}
			\caption[]{Comparison of the combination of various neural networks for features extraction with \gls{UMAP}. The red asterisk represents the position of the same data point.}
			\label{fig:umap_with_various_features_vector}
		\end{figure*}

\begin{figure*}[!htb]
			\centering
			\includegraphics[width=0.7\textwidth]{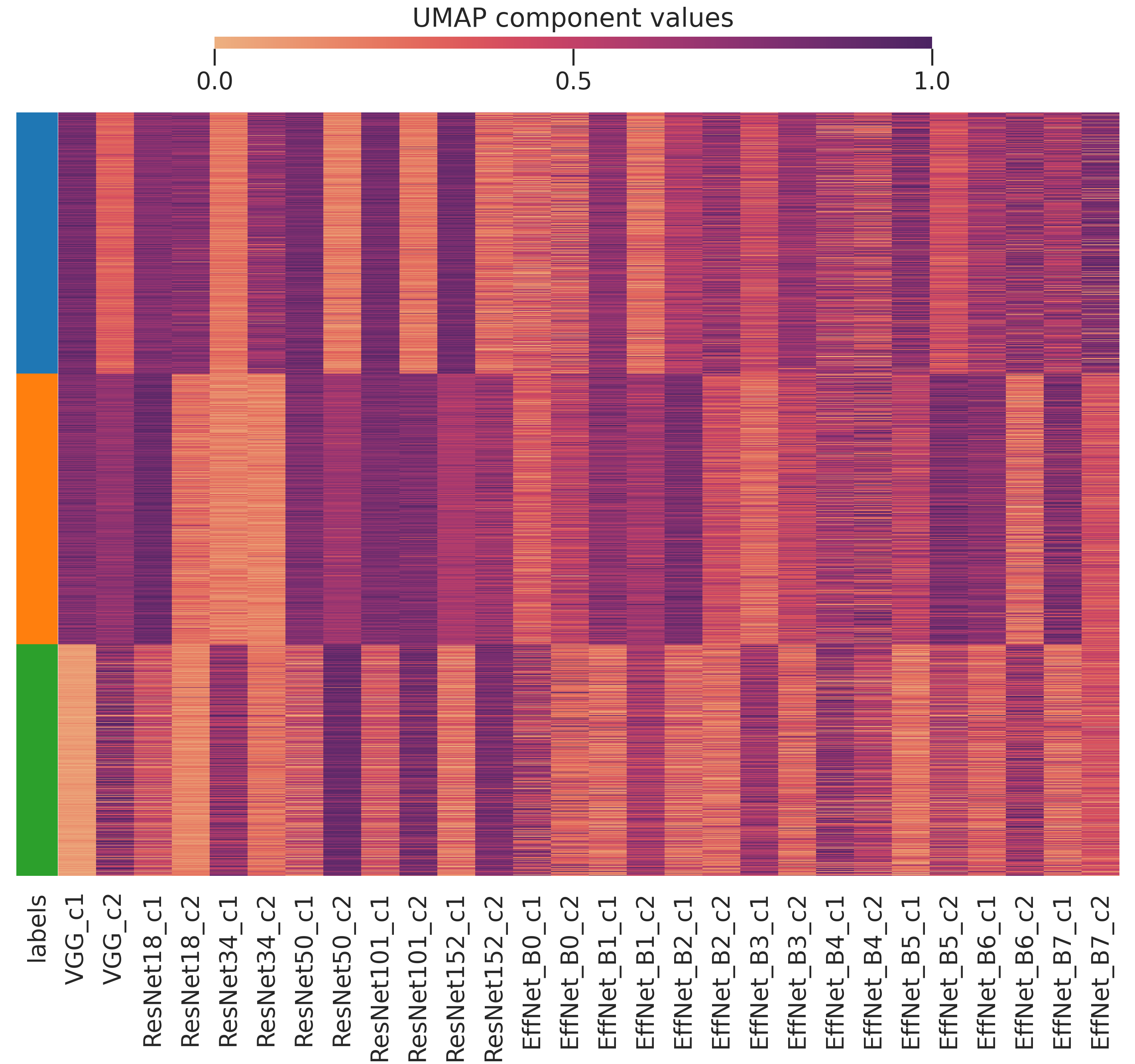}
			\caption[]{Visualization values of the component 1 and component 2 from the combination of various neural networks with \gls{UMAP}. The values along the vertical direction indicate component value. These values are grouped into three dislocation types: \gls{BPD}, \gls{TED} and \gls{TSD} with corresponding colors: blue, orange and green in the left most column.}
			\label{fig:umap_with_various_features_vector_clustermap}
		\end{figure*}

%===============================================================================
In the main text, we have demonstrated that the combination of VGG-16 and \gls{UMAP} 
resulted in a direct neighbor-ship of similar etch-pits which was important 
for obtaining overall very satisfactory results. To evaluate the performance of this combination, we have compared it with $13$ other combinations of various neural networks with \gls{UMAP}. The others neural network consists of ResNet \citep{he2016deep} (with $18$, $34$, $50$, $101$ and $152$ layers) and EfficientNet \citep{tan2019efficientnet} from $B0$ to $B7$. The evaluation is done on $3000$ etch-pits that are selected belonging to \gls{BPD}, \gls{TED} and \gls{TSD}. From the scatter plots and the cluster map plots  shown in \cref{fig:umap_with_various_features_vector} and \cref{fig:umap_with_various_features_vector_clustermap} we can see that \gls{UMAP} groups etch-pits from different dislocation types very effectively if the features vector is obtained from the VGG-16 or the ResNet50, whereas  parts or the whole cluster  overlap for other method combinations. The component values are scaled to the range $0..1$ for comparison purpose.

\begin{figure*}[!htb]
			\centering
			\includegraphics[width=1\textwidth]{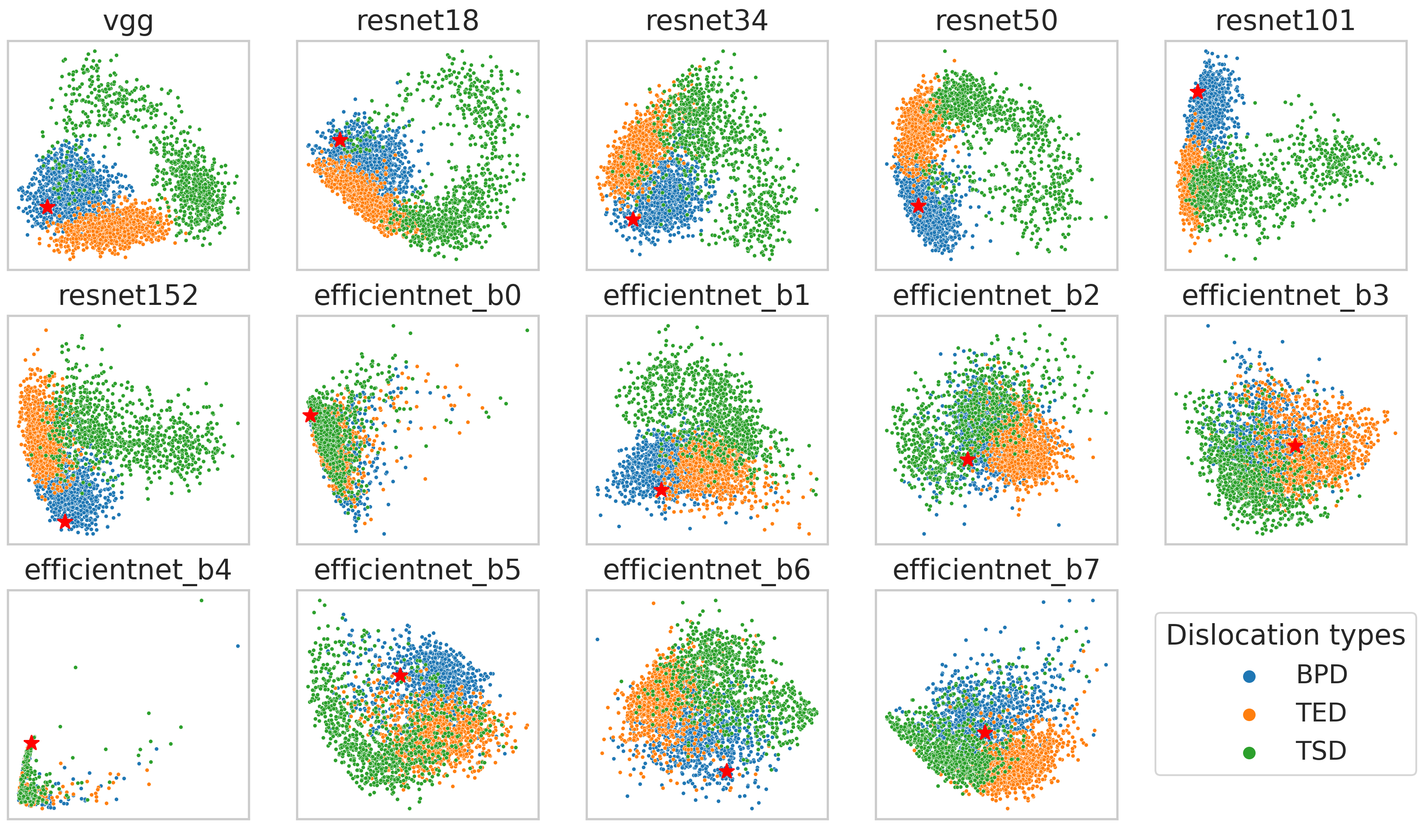}
			\caption[]{Comparison of the combination of various neural network for features extraction with \gls{PCA}. The red asterisk represents the position of the same data point.}
			\label{fig:pca_with_various_features_vector_v2}
		\end{figure*}

\begin{figure*}[!htb]
			\centering
			\includegraphics[width=0.7\textwidth]{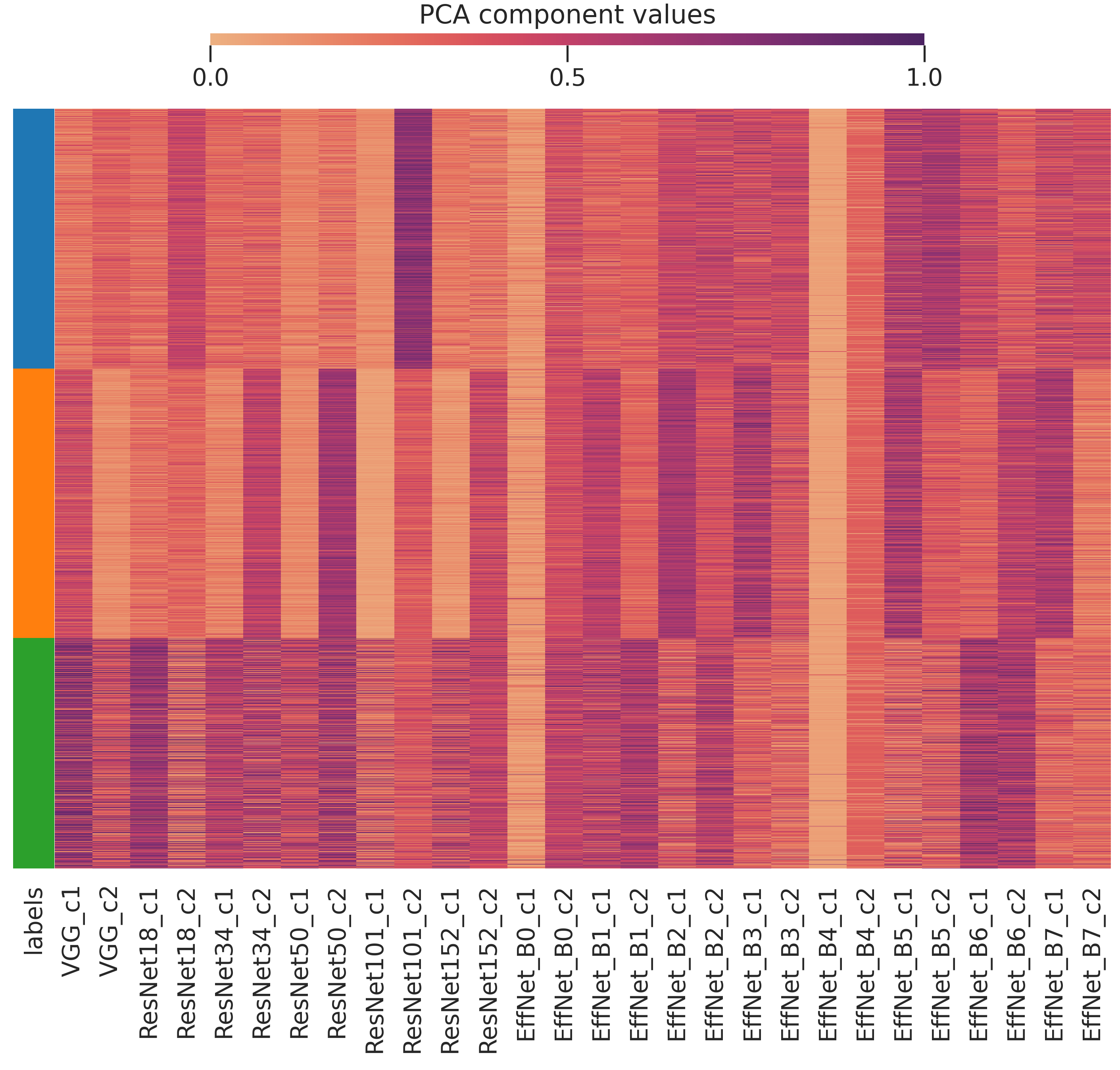}
			\caption[]{Visualization values of the component 1 and component 2 from the combination of various neural networks with \gls{PCA}. The values along the vertical direction indicate component value. These values are grouped into three dislocation types: \gls{BPD}, \gls{TED} and \gls{TSD} with corresponding colors: blue, orange and green in the left most column.}
			\label{fig:pca_with_various_features_vector_clustermap}
		\end{figure*}

\begin{figure*}[!htb]
			\centering
			\includegraphics[width=1\textwidth]{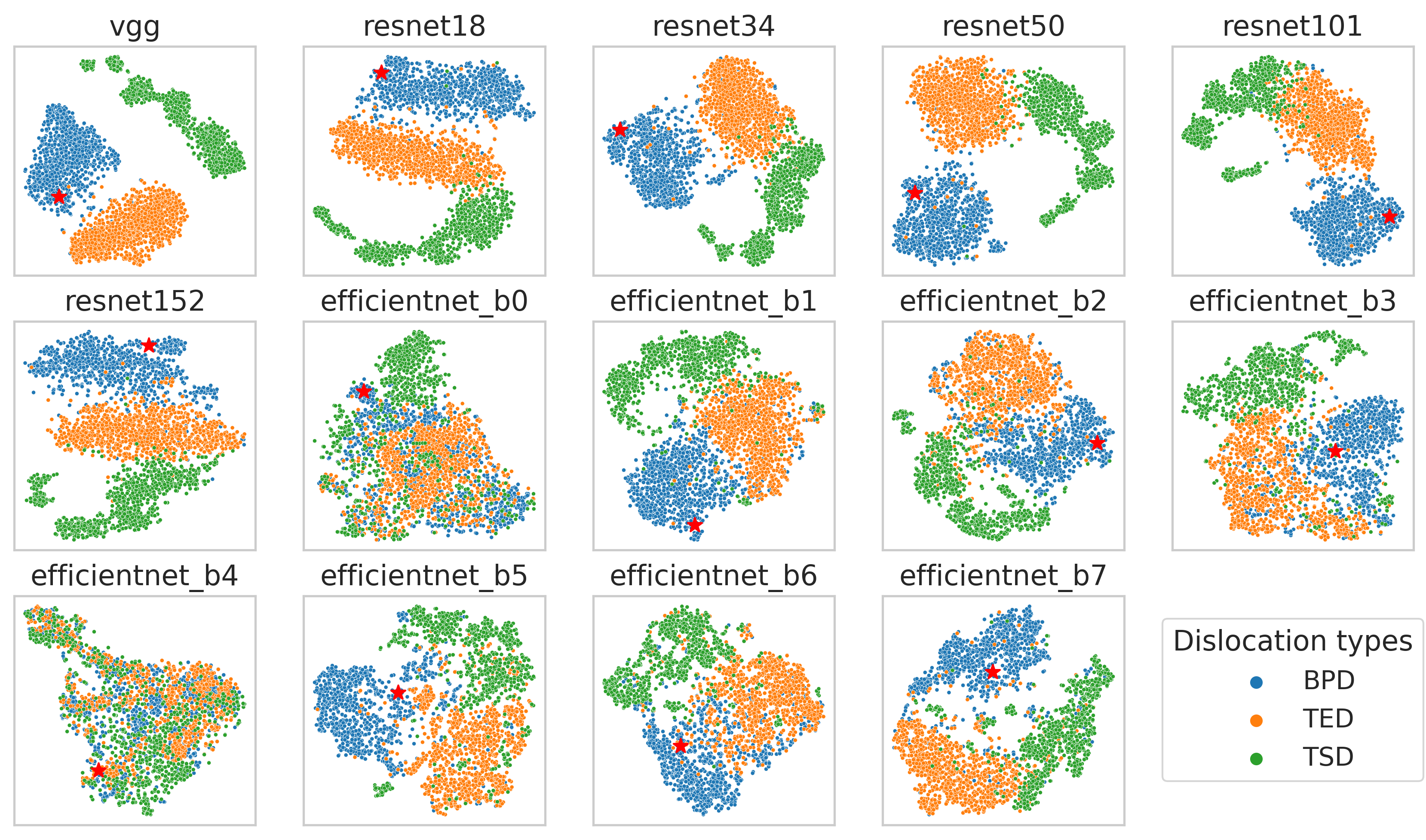}
			\caption[]{Comparison of the combination of various neural network for features extraction with \gls{t-SNE}. The red asterisk represents the position of the same data point.}
			\label{fig:tsne_with_various_features_vector_v2}
		\end{figure*}

\begin{figure*}[!htb]
			\centering
			\includegraphics[width=0.7\textwidth]{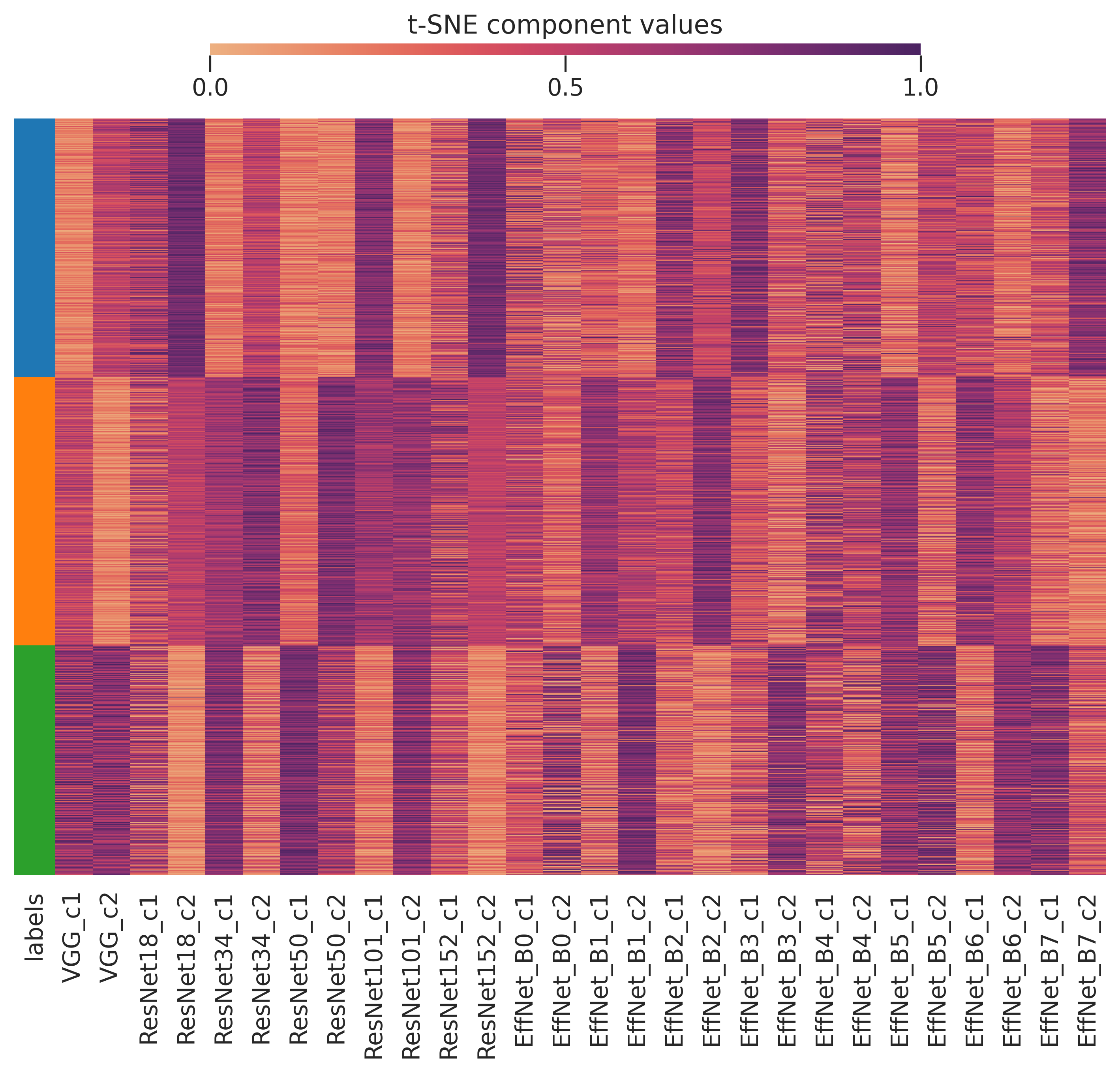}
			\caption[]{Visualization values of the component 1 and component 2 from the combination of various neural networks with \gls{t-SNE}. The values along the vertical direction indicate component value. These values are grouped into three dislocation types: \gls{BPD}, \gls{TED} and \gls{TSD} with corresponding colors: blue, orange and green in the left most column.}
			\label{fig:tsne_with_various_features_vector_clustermap}
		\end{figure*}

Additionally, a similar evaluation was also done using \gls{PCA} (see \cref{fig:pca_with_various_features_vector_v2} and \cref{fig:pca_with_various_features_vector_clustermap}) and \gls{t-SNE} (see \cref{fig:tsne_with_various_features_vector_v2} and \cref{fig:tsne_with_various_features_vector_clustermap}). In \gls{PCA}, although most of the similar etch-pits stay close together, there is no clear separation between pits of different classes. \gls{t-SNE} shows a clear separating boundary between clusters similar to \gls{UMAP} in the some method combinations, such as with VGG-16 or ResNet. Therefore, \gls{UMAP} or \gls{t-SNE} are both suitable methods. However, with larger datasets, the analysis with \gls{UMAP} is recommended because the clustering process requires significantly less computational time \citep{mcinnes2018umap}.

		%=============================================%%
		% For submissions to Nature Portfolio Journals %%
		% please use the heading ``Extended Data''.   %%
		%=============================================%%
		
		%=============================================================%%
		% Sample for another appendix section			       %%
		%=============================================================%%
		
		% \section{Example of another appendix section}\label{secA2}%
		% Appendices may be used for helpful, supporting or essential material that would otherwise 
		% clutter, break up or be distracting to the text. Appendices can consist of sections, figures, 
		% tables and equations etc.
		
	\end{appendices}

    \clearpage 
    \twocolumn
	
	\bibliography{literature}% common bib file
	
	
\end{document}